# TRUTH MAINTENANCE UNDER UNCERTAINTY

Li-Min Fu

Department of Electrical Engineering
and Computer Science
The University of Wisconsin-Milwaukee
Milwaukee, Wisconsin 53201

**abstract**

This paper addresses the problem of resolving errors under uncertainty in a rule-based system. A new approach has been developed that reformulates this problem as a neural-network learning problem. The strength and the fundamental limitations of this approach are explored and discussed. The main result is that neural heuristics can be applied to solve some but not all problems in rule-based systems.

## 1 Introduction

The neural-network approach, also referred to as the connectionist approach or parallel distributed processing, adopts a *brain metaphor* of information processing [9]. Under this approach, information processing occurs through interactions among a large number of simulated neurons each of which is quite limited in its processing capabilities. Descriptions of the properties of connectionist models can be found in [4].

The *tabula rasa* (neural net) approach to machine learning turns out a failure [10]. Even if a neural net can gradually evolve to adapt itself to the external environment, the randomness of its learning behavior makes the process of evolution extremely slow. It has long been argued that a close resemblance between the computer's internal representations and neural nets is neither necessary nor feasible. Neural networks implemented earlier are often severly limited in the kinds of computations they can perform. However, recent successes of the neural-network approach demonstrated in solving such problems as learning to speak, recognizing hand-written characters, and signal processing have forced a reconsideration of this approach to the production of intelligent behavior. The resurgence of the interest in this approach is partly a consequence of recent hardware advances in the construction of massively parallel computers that enable much faster simulation of neural networks [9]. Yet, we should not be over-optimistic about this approach because neural networks implemented nowadays are several orders of magnitude smaller than those observed in biological systems. So far, it has not yet been demonstrated that the neural-network approach can learn high level knowledge.

Several successful knowledge-based systems have been built since the middle seventies, such as DENDRAL, XCON, PROSPECTOR, and CADUCEUS. In artificial intelligence research, it has been recognized that to learn new knowledge a computer program must possess a great deal of initial knowledge. Since, however, human intelligent behavior seems to emerge from interactions among a huge number of neurons, the knowledge-based approach does not capture this fundamental nature of intelligence as much as the neural-network approach.

The neural-network approach contrasts with the knowledge-based approach in several aspects. The knowledge of a neural network lies in its connections and associated weights, whereas the knowledge of a rule-based system lies in rules. A neural network processes information by propagating and combining activations through the network, but a knowledge-based system reasons through symbol generation and pattern matching. The knowledge-based approach emphasizes knowledge representation, reasoning strategies, and the ability to explain, whereas the neural-network approach does not. The knowledge-based approach can reason at various levels of abstraction, but the neural-network approach cannot. The key differences between these two

Table 1: Comparison between the neural-network and the knowledge-based approaches

|  | the neural-network approach | the knowledge-based approach |
|---|---|---|
| Knowledge | connections | rules |
| Computation | numbers | numbers, symbols |
|  | summation and thresholding | pattern matching |
|  | simple, uniform | complicated, various |
| Reasoning | non-strategic | strategic, meta-level |
| Tasks | signal level | knowledge level |

approaches are summarized in table 1.

The confluence of recent successes of these two approaches, the neural-network and the knowledge-based approaches, leads one to think that combining the technologies of these two approaches is promising. We can envision several possible ways to combine these two approaches:

- Applying neural heuristics (heuristics developed under the neural-network approach) to solve problems in building knowledge-based systems.

- Adding knowledge to computer-based neural-networks.

- Developing hybrid approaches which combine the desirable features of these two paradigms.

- Integrating these two approaches in one system.

However, the approach which employs connectionist models to solve expert problems (see [7], for example) is not included in the above list since this approach relies solely on the neural technology. Among these alternatives, only the third one has been limitedly explored, see [5], for example. In this paper, we will focus on the first alternative for combining these two approaches.

The remaining sections are organized as follows. Section 2 formulates the addressed problem, namely the problem of error handling in rule-based systems. Section 3 describes how to reformulate this problem as a neural-network learning problem and discusses the limitations of this reformulation. Section 4 presents the approach which applies the *back-propagation* heuristic to the addressed problem. This section addresses the following topics: representing the knowledge base and input data as a coherent network, back-propagating the errors of belief values, distinguishing knowledge errors from data errors, and debugging the knowledge base and input data. Section 5 describes the evaluation of the developed approach. The last section draws conclusions about this approach.

## 2 The Problem

When errors are observed in the conclusions made by a rule-based system, an issue is raised of how to identify and correct the rules or data responsible for these errors. The problem of identifying the sources of errors is known as the *blame assignment* problem.

Previous approaches including [2 11 13 14] only focus on how to revise the knowledge base. Among these, TEIRESIAS [2] is the typical work. It maintains the integrity of the knowledge base by interacting with experts. However, as the size of the knowledge base grows, it is no longer feasible for human experts to consider all possible interactions among knowledge in a coherent and consistent way. TMS [3] resolves inconsistency by altering a minimal set of beliefs, but it lacks the notion of uncertainty in the method itself.

Conventional approaches such as TEIRESIAS suffer from several problems. First, they do not consider the case where incorrect conclusions are due to data errors. Second, they assume that the strengths of rules involving intermediate concepts can be determined by experts. In fact, expert knowledge for the most part concerns the associations between observable data and final hypotheses. Third, they lack a consistent and coherent means for modifying the strengths of rules. Experts may know the strength of inference of each individual rule but it may be difficult for them to determine the rule strengths in a way such that dependencies among rules are carefully considered in order to meet the system's assumptions. For instance, in MYCIN-like systems [1], since certainty factors are combined under the assumption of independence, the certainty factors assigned to two dependent rules should be properly adjusted so as to meet this assumption. The approach developed in this paper will address all the three problems.

## 3 Reformulation of The Problem

A rule-based system (knowledge represented in rules) can be transformed into an infer-ence network where each connection corresponds to a rule, and each node corresponds the premise or the conclusion of a rule. Reasoning in such systems is a process of propagating and combining multiple pieces of evidence through the inference network until final conclusions are reached. Uncertainty is often handled by adopting the certainty factor (CF) or the probabilistic schemes which associate each fact with a number called *belief value*. An important part of reasoning tasks is to determine the belief values of predefined final hypotheses given the belief values of observed evidence. If we deprive an inference system of all properties except belief values, the skeleton left is a *belief network*. Correspondence of structural and behavioral aspects exists between neural networks and belief networks, as shown in table 2. For instance, the summation function in neural networks corresponds to the function for combining certainty factors in MYCIN-like systems or to the Bayesian formula for deriving *a posteriori* probabilities from combined evidences in PROSPECTOR-like systems. The thresholding function in neural networks corresponds to predicates such as SAME in MYCIN-like systems which cuts off any certainty value below .2. Since belief networks correspond to neural networks in every structural and behavioral attribute shown in table 2, any algorithm that is applicable to neural networks characterized by no more than these attributes may also be applicable to belief networks. The *back-propagation* rule is just such an algorithm.

Table 2: Correspondence between neural networks and belief networks

| Neural networks | Belief networks |
|---|---|
| connections | rules |
| nodes | premises, conclusions |
| weights | rule strengths |
| thresholds | predicates |
| summation | combination of belief values |
| propagation of activations | propagation of belief values |

From the above analyses, the problem of error handling in rule-based systems can be reformulated as a neural-network learning problem and solved accordingly. However, errors that involve aspects other than belief values cannot be solved this way.

## 4 Truth Maintenance

In this section, a neural-network approach to handling errors in rule-based systems is presented. A technique of representing the knowledge base and input data as a coherent network is introduced. The application of the *back-propagation* rule to rule-based systems is examined. A method of distinguishing knowledge errors from data errors is described. Finally, revision techniques under this approach are analyzed.

### 4.1 The Network of the Knowledge Base and Input Data

A knowledge base represented in rules can be structured as a network, with each node representing a fact and each connection representing a rule. The starting and the ending nodes associated with a connection represent respectively the premise and the conclusion parts of the corresponding rule. The weight of a connection is just the strength of the corresponding rule. In this network, output nodes (nodes without connections pointing outwards) stand for final hypotheses and input nodes (nodes without connections pointing inwards) stand for data attributes. The remaining nodes are intermediate nodes, which represent concepts that summarize or categorize subsets of data or intermediate hypotheses that infer final hypotheses.

If there are no data errors, the input nodes of the knowledge base network can represent both the observed and the actual inputs. In case of possible data errors, the observed input and the actual input are represented as two different levels of nodes, with a connection established between each observed and actual input nodes referring to the same data attribute. One example is shown in figure 1 where, for instance, observed data node $E'_1$ corresponds to actual data node $E_1$.

In this way, the knowledge base and input data are organized as a coherent network so that whenever an error arises, it can be propagated to responsible loci without making distinctions between knowledge and data. Thus, they can be revised consistently.

### 4.2 Back-Propagation of Error

An error refers to the disagreement between the belief value generated by the system and that indicated by a knowledge source assumed to be correct (e.g., an expert) with respect to some fact. The *back-propagation* rule developed in the neural network approach [12] is a recursive heuristic which propagates backwards errors at a node to all nodes pointing to that node, and modifies the weights of connections leading into nodes with errors.

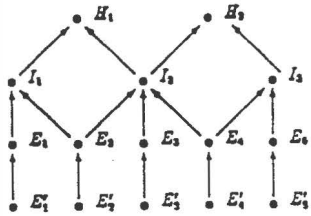

Figure 1: Organization of the knowledge base and input data as a network.

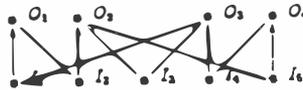

Figure 2: A single-layered network.

First, we restrict our attention to single-layered networks involving only input nodes and output nodes such as the one shown in figure 2.

In each inference task, the system arrives at the belief values of final hypotheses given those of input data. The belief values of input data form an input pattern (or an input vector) and those of final hypotheses form an output pattern (or an output vector). Systems errors refer to the case when incorrect output patterns are generated by the system. When a system's error arises in an inference task, we use the instance consisting of the input pattern and the correct output pattern to train the system. The instance is repeatedly used to train the network until a satisfactory performance is reached. Since the network may be incorrectly trained by that instance, we also maintain a set of reference instances to monitor the training process. This reference set is consistent with the knowledge base. If, during training, some instances in the reference set become inconsistent, they will be added to the training.

On a given trial, the network generates an output vector given the input vector of the training instance. The discrepancy obtained by subtracting the network's from the desired output vectors serves as the basis for adjusting the strengths of the connections involved. The *back-propagation* rule adapted from [12] is formulated as follows:

$$\triangle W_{ji} = r D_j (dO_j/dW_{ji}) \qquad (1)$$

where

$$D_j = T_j - O_j,$$

$\triangle W_{ji}$ is the weight (strength) adjustment of the connection from input node $i$ to output node $j$, $r$ is a trial-independent learning rate, $D_j$ is the discrepancy between the desired belief value ($T_j$) and the network's belief value ($O_j$) at node $j$, and the term $dO_j/dW_{ji}$ is the derivative of $O_j$ with respect to $W_{ji}$. According to this rule, the magnitude of weight adjustment is proportional to the product of the discrepancy and the derivative above.

**Theorem 1.** The *back-propagation* rule is applicable to belief networks where the propagation and the combination of belief values are determined by differentiable mathematical functions.

**Proof.** As shown in Eq.(1), the mathematical requirement of the *back-propagation* rule is that the relationship between the output activation ($O_j$) and the input weight ($W_{ji}$) is determined by a differentiable function. In belief networks, this relationship is differentiable if the propagation and the combination functions are differentiable. □

Since combining belief values in most rule-based systems involves such logic operations as conjunction or disjunction, the *back-propagation* rule cannot be applied without using some trick. The trick we employ turns the conjunction operator into multiplication and turns the disjunction operator into summation.

We are now in a position to examine multi-layered networks. A multi-layered network, as depicted in figure 1, involves at least three levels: one level of input nodes, one level of output nodes, and one or more levels of middle nodes.

Learning in a multi-layered network is more difficult because the behavior of middle nodes is not directly observable. Modifying the strengths of the connections pointing to a middle node entails the knowledge of the discrepancy between the network's and the desired belief values at the middle node. The discrepancy at a middle node can be derived from the discrepancies at output nodes which receive activation from the middle node [12]. Accordingly, the discrepancy at middle node $j$ is defined by

$$D_j = \sum_k W_{kj} D_k$$

where $D_k$ is the discrepancy at node $k$. In the summation, each discrepancy $D_k$ is weighted by the strength of the connection pointing from middle node $j$ to node $k$. This is a recursive definition in which the discrepancy at a middle node is always derived from discrepancies at nodes at the next higher level.

In addition, the belief value of a middle node can be obtained by propagating the belief values at input nodes recursively and combining these values properly until the middle node is reached.

### 4.3 Distinguishing Knowledge-Base Errors from Input Data Errors

We devise a method that can distinguish knowledge-base errors from input data errors. This method includes three tests. In the first test, we clamp all connections corresponding to the knowledge base so that only the strengths of the connections between the observed and the actual input data nodes remain adjustable during training. In the second test, we clamp the connections between the observed and the actual inputs and allow only the strengths of the connections corresponding to the knowledge base to be modified. In the third test, we allow the strengths of all connections to be adjusted. In each test, success is reported if the error concerned can be resolved after training; failure is reported otherwise. As a result, there are eight possible outcomes combined from the results of these three tests.

| Test 1 | Test 2 | Test 3 | Outcome |
|--------|--------|--------|---------|
| Success | Success | Success | $O1$ |
| Success | Success | Failure | $O2$ |
| Success | Failure | Success | $O3$ |
| Success | Failure | Failure | $O4$ |
| Failure | Success | Success | $O5$ |
| Failure | Success | Failure | $O6$ |
| Failure | Failure | Success | $O7$ |
| Failure | Failure | Failure | $O8$ |

These results are interpreted as follows. Outcome $O1$ suggests the revision of either the knowledge base or input data. In this case, we need an expert's opinion to decide which should be revised. Outcome $O2$ is unlikely in our experience and is ignored. Outcome $O3$ suggests the revision of input data. Outcome $O4$ is unlikely and is ignored. Outcome $O5$ suggests the revision of the knowledge base. Outcome $O6$ is also unlikely and is ignored. Outcome $O7$ suggests the revision of both the knowledge base and input data. Outcome $O8$ is a deadlock, which demands an expert to resolve the error.

### 4.4 Revision Operations

The results of the above tests will indicate whether the knowledge base or input data or both should be revised. The strengths of the connections in the network (representing the knowledge base and input data) have been revised after training. The next question is how to revise the knowledge base and/or input data according to the revisions made in the network. We will first focus on the revision of the knowledge base.

Basically, there are five operators for rule revision: *modification of strengths, deletion, generalization, specialization,* and *creation* [2]. However, not all the five operators are suitable in the neural-network approach to editing rules. We will examine each operator.

The *modification of strengths* operator is not necessary since the strength of a rule is just a copy of the weight of the corresponding connection and the weights of connections have been modified after training with the *back-propagation* rule.

The *deletion* operator is justified by theorem 2.

**Theorem 2.** In a rule-based system, if the following conditions are met:

1. the belief value of the conclusion is determined by the product of the belief value of the premise and the rule strength,

2. the absolute value of any belief value and rule strength is not greater than 1,

3. any belief value is rounded off to zero if its absolute value is below threshold $k$ ($k$ is a real number between 0 and 1), and

then the deletion of rules with strengths below $k$ will not affect the belief values of the conclusions arrived at by the system.

**Proof.** From conditions 1 and 2, if the strength of rule $R$ is below $k$, the belief value of its conclusion is always below $k$. From conditions 3, the belief value of the conclusion made by rule $R$ will always be rounded off to zero. Since rule $R$ is not effective in making any conclusion, it can be deleted. Thus, the deletion of such rules as rule $R$ will not affect the system's conclusions. □

Accordingly, deletion of a rule is indicated when its absolute strength is below a predetermined threshold. In MYCIN-like systems, the threshold is .2.

Generalization of a rule can be done by removing some conditions from its premise, whereas specialization can be done by adding more conditions to the premise. If the desired belief value of a conclusion is always higher than that generated by the network and the discrepancy is resistant to decline during training, it is suggested that the rules supporting this conclusion are generalized. On the other hand, if the discrepancy is negative and resistant, specialization is suggested. However, generalization or specialization of a rule may involve qualitative changes of a node. The *back-propagation* rule has not yet been powerful enough to make this kind of changes.

Creation of new rules involves establishment of new connections. Just as we delete a rule if its absolute strength is below a threshold, we may establish a new connection when its absolute strength is above the threshold. To create new rules, we need to create some additional connections which can potentially become rules. Without any bias, one may need an inference network where all data are fully connected to all intermediate hypotheses, which in turn are fully connected to all final hypotheses. This is not a feasible approach unless the system is small.

From the above analyses, we allow only the *modification of strengths* operator and the *deletion* operator in the neural-network approach to rule revision.

Revision of input data is much simpler. If the strength of the connection between an observed and an actual input nodes is below a predetermined threshold, the corresponding input data attribute is treated as false and deleted accordingly.

## 5 Evaluation

In this section, we will first demonstrate the developed approach in a practical domain, namely, the problem of diagnosing jaundice, then compare this approach with conventioal approaches such as TEIRESIAS, and finally discuss the applicability of this approach to a large rule-based system with thousands of rules.

Derived from JAUNDICE [6], a rule base contains 50 rules, 5 final hypotheses, 3 intermediate hypotheses, and 20 clinical attributes. This rule base is mapped into a bi-layered network with 5 output nodes, 3 middle nodes, and 20 input nodes. Twenty training instances that can be diagnosed correctly by these 50 rules are collected from the JAUNDICE case library.

Ten experiments were carried out. In each experiment, a small number of incorrect connections (rules) that contradict medical knowledge are added to the network described above. These incorrect rules are provided by a medical expert. No any incorrect rule is shared by two experiments. Then the rule base is used to diagnose the 20 training instances before and after it is revised under the developed approach. The objective of these experiments is to see whether those incorrect rules can be removed. In each experiment, we record the number of incorrect rules and the diagnostic accuracy before and after the error handling procedure is applied to the rule base. The results are shown in table 3. We use the statistical *paired t test* to judge whether the procedure can remove incorrect rules and improve the system's performance significantly. Two *null hypotheses* are formulated. The first states that there is no difference of incorrect rule numbers between before and after the procedure is applied. The second states that there is no difference of the system's performance between before and after the procedure is used. The $t$ values for the first and the second hypotheses are $t=6.32$ and $t=5.85$ respectively. Both hypotheses are rejected at level of significance $\alpha < .01$. In other words, this approach is effective in our experiments. Another result is that no any rule among the original 50 correct rules is deleted.

One important question is whether this approach can be scaled up to the order of rule bases large enough for industrial application such as XCON, which contains thousands of rules. The 50 rule expert system used in the above experiments is certainly too small. However, our claim that the neural-network approach is applicable to error handling in rule-based systems is mainly derived from the following bases. First, we show earlier the analogy between neural networks and belief networks. Second, we show the applicability of the *back-propagation* rule in belief networks (theorem 1). Third, we justify the *deletion* operator of this approach (theorem 2). The results of the experiments can serve as a piece of evidence supporting the claim. Since the problem is solved by reformulating it as a neural-network learning problem, we can envision that the size of the largest rule bases to which the back-propagation rule can be successfully applied is just the size of the largest computer-based neural-networks implementable. Currently, the largest neural networks for real-world application contains several hundred nodes and about ten thousand connections [8]. Thus, in theory, the neural-network approach is applicable to handling errors of rule bases with up to about ten thousand rules.

Table 3: Deletion of incorrect rules and improvement of Diagnostic Accuracy. Before: before revision, After: after revision.

| Experiment | Bad Rule Number | | Diagnostic Accuracy | | Improvement |
|---|---|---|---|---|---|
| | Before | After | Before | After | |
| 1 | 1 | 0 | 100% | 100% | 0% |
| 2 | 2 | 0 | 90% | 100% | 10% |
| 3 | 3 | 0 | 90% | 100% | 10% |
| 4 | 4 | 1 | 80% | 95% | 15% |
| 5 | 5 | 1 | 80% | 95% | 15% |
| 6 | 6 | 1 | 80% | 95% | 15% |
| 7 | 7 | 2 | 70% | 95% | 25% |
| 8 | 8 | 2 | 70% | 95% | 25% |
| 9 | 9 | 2 | 65% | 90% | 25% |
| 10 | 10 | 2 | 60% | 90% | 30% |
| Average | 5.5 | .9 | 78.5% | 95.5% | 17% |

Table 4: Comparison between TEIRESIAS and the neural-network approach

| | TEIRESIAS | The neural-network approach |
|---|---|---|
| Approach | human experts | back-propagation |
| Operators | modifying strengths, deletion, addition, generalization, specialization | modifying strengths, deletion |
| Errors | rule errors | rule and data errors |

It has been known that noise associated with training instances will affect the quality of learning. In the neural-network approach, since noise will be distributed over the network, its effect on individual connections is relatively minor. In practice, perfect training instances are neither feasible nor necessary. As long as most instances are correct, a satisfactory performance can be achieved. Training instances are usually obtained from one or more of the following sources: experience, literature and textbooks, and experts.

The comparison between the TEIRESIAS approach and the neural-network approach to error handling is shown in table 4. The neural-network approach may be more useful than TEIRESIAS in handling multiple errors or errors involving some unobservable concepts which human experts may have difficulties in dealing with. In addition, the back-propagation rule can be uniformly applied to the whole rule base, whereas human experts may focus on certain parts of the rule base consciously or subconsciously. Also, in [14], it is suggested that the only proper way to cope with deleterious interactions among rules is to delete offending rules. In light of this view, the *deletion* operator could be a useful operator. While the neural-network approach is still too simple to deal with errors involving qualitative changes of rules, reasoning strategies, and meta-level knowledge, the techniques developed under this approach can supplement the current rule base technology.

## 6 Conclusion

It has long been argued that the neural-network approach to artificial intelligence is neither necessary nor feasible. However, recent successes of the neural-network approach demonstrated in solving such problems as learning to speak, recognizing hand-written characters, and signal processing have forced a reconsideration of this approach to the production of intelligent behavior. Another approach, the knowledge-based approach, emphasizes that knowledge is the key to the generation of intelligent computer systems. Since the middle seventies, a number of successful knowledge-based systems have been built. It is natural to ask which approach is more promising. Instead of answering this question, we begin to explore the approach which combines the technologies of these two approaches.

In this work, a new approach has been developed that applies neural heuristics to the problem of error handling in rule-based systems. This approach reformulates the problem as a neural-network learning problem. The rationale behind this reformulation is the analogy between neural networks and belief networks. The techniques developed under this approach identify and revise rules responsible for system's errors by employing the *back-propagation* rule.

We have demonstrated in our experiments that this approach can effectively delete incorrect rules and improve the system's performance. According to our analyses, this ap-proach can be scaled up to the order of rule bases with thousands of rules, though we need more experience to confirm this claim.

The capabilities of this approach are limited to modifying rule strengths and deletion of incorrect rules. However, the deletion of rules could be the only proper way to cope with deleterious interactions among rules [14]. It is discussed that this new approach may be more useful than conventional approaches such as TEIRESIAS in case of multiple errors, errors involving hidden concepts, or data errors. Although, currently, the techniques developed under this approach are limited in the kinds of operations they can perform, our research indicates that these techniques can supplement the rule base technology.